\documentclass[11pt]{article}

\usepackage{amsmath}
\usepackage{amssymb}

\usepackage[preprint]{acl}

\usepackage{times}
\usepackage{latexsym}

\usepackage[T1]{fontenc}

\usepackage[utf8]{inputenc}

\usepackage{microtype}

\usepackage{inconsolata}

\usepackage{graphicx}
\usepackage{booktabs}
\usepackage{colortbl}
\usepackage{multirow}
\definecolor{gapblue}{rgb}{0.88,0.95,1}
\newcommand{\gapcell}[1]{\cellcolor{gapblue}#1}
\definecolor{predzero}{RGB}{168,200,230}
\newcommand{\zerocell}[1]{\cellcolor{predzero}#1}

\title{Readable Yet Unpredictable: Rotated-Outcome Prediction in Vision-Language Models}

\author{
  \textbf{Lexin Wang\textsuperscript{1}},
  \textbf{Shenghua Liu\textsuperscript{1}\thanks{Corresponding author.}},
  \textbf{Yiwei Wang\textsuperscript{2}},
  \textbf{Jiafeng Guo\textsuperscript{1}},
  \textbf{Xueqi Cheng\textsuperscript{1}}
\\
\\
  \textsuperscript{1}Institute of Computing Technology, Chinese Academy of Sciences
\\
  \textsuperscript{2}University of California, Merced
\\
  \url{https://rotoutbench.netlify.app/}
}

\begin{document}
\ifdefined\linenumbers\linenumbers\fi
\maketitle
\begin{abstract}
Can vision-language models predict what a 180$^\circ$ rotation would reveal from the original image alone? We study this ability through \textit{Rotated-Outcome Prediction}: given an original image, a model must answer what would be seen or read after a 180$^\circ$ in-plane rotation, without directly observing the rotated target. To isolate this gap, we introduce \textsc{RotOutBench}, a paired diagnostic benchmark spanning open visual cases and controlled text-image rotations. A sharp pattern emerges: many VLMs can recognize the relevant content when directly given either the original or rotated image, yet fail to infer the rotated result from the original image alone. On controlled text-image rotations, predicted-rotation accuracy collapses to near zero even for models with high direct-reading accuracy. A model-level diagnostic further shows that the prediction state can approach a rotated-image reading state, while the final readout still shifts toward the original string. Current VLMs can recognize a transformed visual state when it is shown, but often fail to predict that state from the original view.
\end{abstract}

\begin{figure*}[t]
\centering
\includegraphics[width=\textwidth]{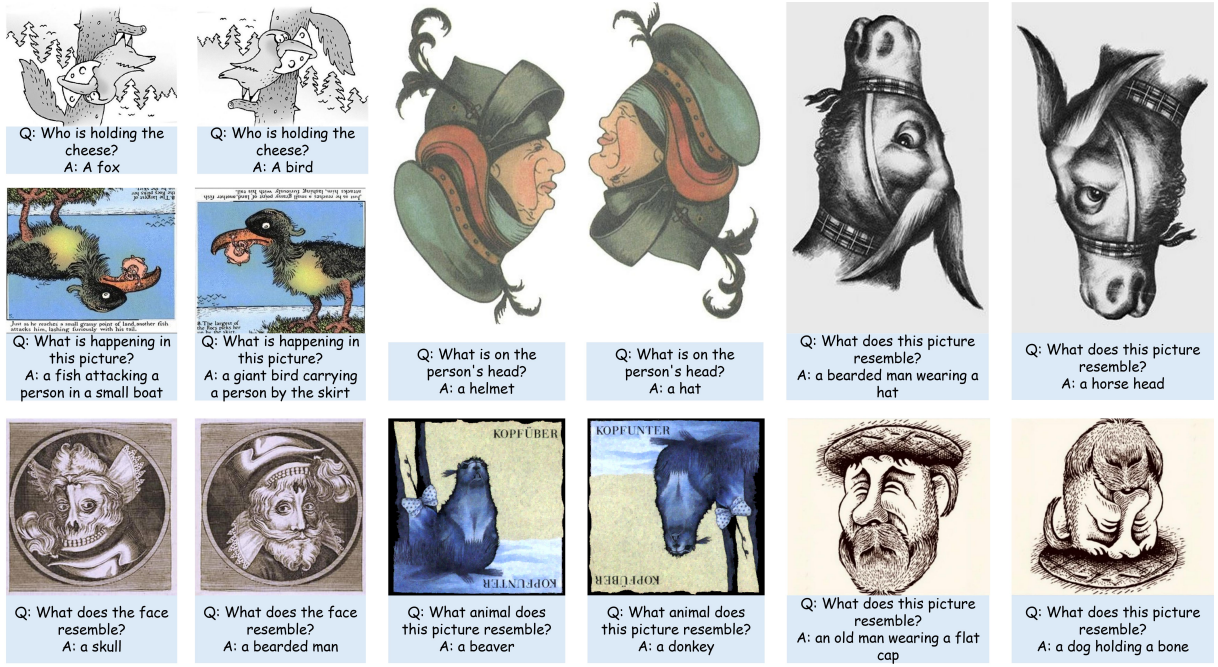}
\caption{Examples from Visual-Rot. The benchmark includes images whose interpretation can change under a 180$^\circ$ rotation, requiring models to predict an unobserved rotated visual outcome rather than only recognize the currently visible image.}
\label{fig:visual_rot_examples}
\end{figure*}

\section{Introduction}

Vision-language models (VLMs)~\citep{radford2021learning,li2022blip,alayrac2022flamingo,liu2023visualinstruction} have achieved strong performance on image understanding and multimodal reasoning tasks. Yet visual inputs are not always observed in a canonical orientation: documents may be scanned upside down, text may appear at unusual angles, and agents may view objects from non-standard viewpoints. In such cases, recognizing the current image is not always enough; the model may need to infer how the content would appear after a change of orientation.

Existing benchmarks study important parts of this problem. OCR and document-understanding benchmarks evaluate whether models can read visible text~\citep{liu2024ocrbench,fu2026ocrbench}, while spatial, orientation, and rotation benchmarks test whether models can judge relations, viewpoints, or rotated inputs that are directly presented~\citep{stogiannidis2025mind,jung2025right,niu2026rotbench}. These evaluations show that orientation and spatial structure remain challenging for VLMs. However, they mainly evaluate recognition or judgment over visual states that are already shown to the model.

This leaves an under-explored distinction: recognizing a transformed visual state when it is shown is not the same as predicting that state from an observed starting point. We study this ability through \textit{Rotated-Outcome Prediction}. Given an original image, the model must answer what would be seen or read after a 180$^\circ$ in-plane rotation, without directly observing the rotated target. This setting tests whether the model can infer an unseen rotated outcome from the original image alone.

To study this ability, we introduce \textsc{RotOutBench}, a paired diagnostic benchmark with two complementary subsets. Visual-Rot contains 68 paired examples, yielding 136 image views, where natural and icon-like images may change interpretation after rotation. TextImage-Rot contains 342 paired string examples, yielding 684 image views, where each sample has an original rendered string image and a canonical rotated target image. \textsc{RotOutBench} is designed as a focused diagnostic benchmark rather than a large-scale leaderboard: its paired design prioritizes controlled comparison between endpoint recognition and unseen outcome prediction. Figure~\ref{fig:visual_rot_examples} shows representative examples from the Visual-Rot subset.

A striking pattern emerges on TextImage-Rot: many recognition-capable VLMs can directly read the relevant content when given either the original or rotated image, yet their predicted-rotation accuracy collapses to near zero when asked to infer the rotated result from the original image alone. This is not simply a failure to recognize rotated images. Instead, it reveals an endpoint-recognition vs. outcome-prediction disconnect: current VLMs can recognize visible transformed states, but do not reliably derive those states from observed inputs. Four-way rotation matching on Visual-Rot further remains close to the random baseline, suggesting that explicitly presenting candidate images does not make the rotation correspondence reliable.

We further probe this failure with text-only controls and model-level diagnostics. The analysis suggests that the prediction state can approach a rotated-image reading state, but the final readout still favors the original string.

Our contributions are threefold:
\begin{itemize}
    \item We formalize \textit{Rotated-Outcome Prediction}, a diagnostic setting for predicting an unseen rotated outcome from an observed image.
    \item We introduce \textsc{RotOutBench}, a paired benchmark with open visual cases and controlled text-image rotations.
    \item We show that VLMs can recognize transformed endpoints when shown, yet fail to predict the rotated outcome from the original view; model-level diagnostics further reveal that the final readout can favor the original string even when the prediction state approaches the rotated target.
\end{itemize}

\begin{figure*}[t]
\centering
\includegraphics[width=\textwidth]{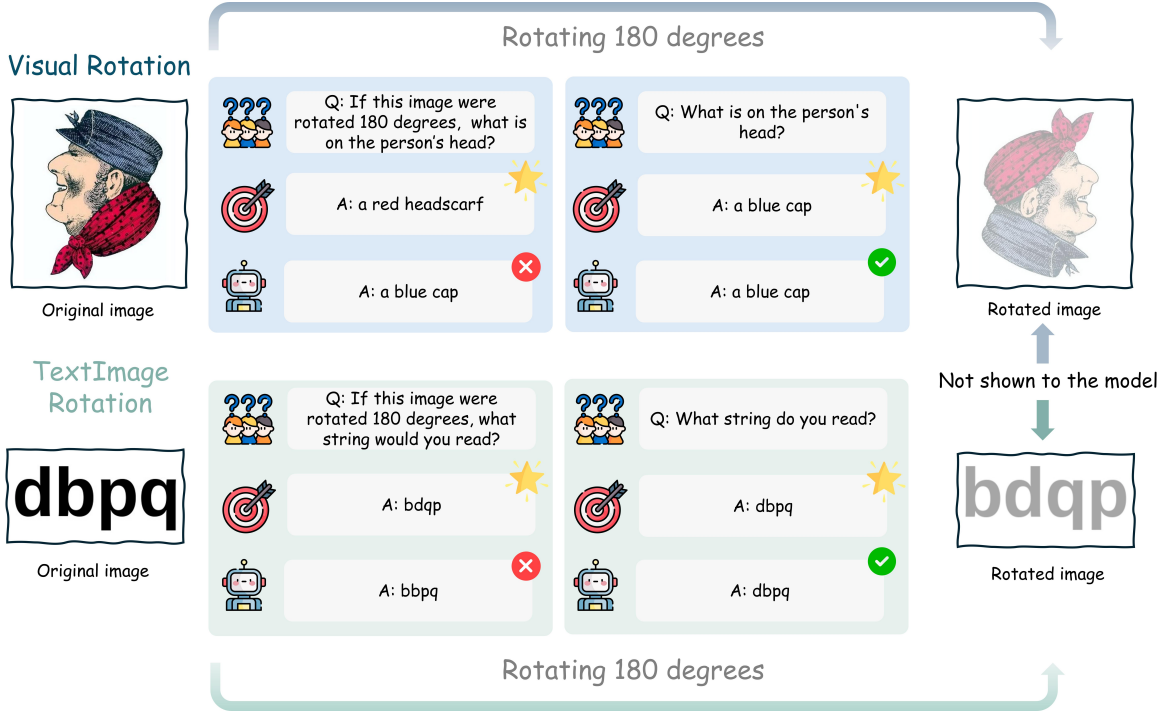}
\caption{Example cases from Visual-Rot and TextImage-Rot.
Each example pairs an original view with its canonical 180$^\circ$ rotated view.
The examples illustrate the gap between directly reading a visible rotated target and predicting the same rotated outcome from the original view.}
\label{fig:benchmark_answer_examples}
\end{figure*}
\section{Related Work}

\subsection{Orientation and Spatial Reasoning in VLMs}

Spatial reasoning has long been used to diagnose visual understanding, from compositional datasets such as CLEVR~\citep{johnson2017clevr} and GQA~\citep{hudson2019gqa} to spatial-relation benchmarks such as VSR~\citep{liu2023visual} and What'sUp~\citep{kamath2023s}. Recent diagnostic suites further show that VLMs may perform well on general VQA while still failing on fine-grained spatial, geometric, and viewpoint-sensitive tasks~\citep{tong2024eyes,fu2024blink,li2024seed,wang2024picture,stogiannidis2025mind}. Visual text and document-understanding benchmarks further show that reading visible text remains an important capability for multimodal models~\citep{liu2024ocrbench,fu2026ocrbench}. These failures suggest that VLMs do not always represent visual text, relations, orientation, and viewpoint-dependent structure robustly.

More closely related are benchmarks on orientation, reference frames, and mental rotation. EgoOrientBench~\citep{jung2025right} and perspective-aware reasoning tasks~\citep{lee2025perspective,zhang2025spinbench} show that models can struggle with egocentric directions and viewpoint changes. RotBench~\citep{niu2026rotbench} evaluates whether MLLMs can identify image orientations, while SpinBench~\citep{zhang2025spinbench} studies rotation and perspective taking as spatial-reasoning probes. These works test orientation recognition, comparison, or perspective reasoning when the relevant visual inputs or candidate views are provided. Our setting differs in that the rotated image is not shown: the model must predict the rotated visual-semantic outcome from the original image.

\subsection{Visual Grounding, Ambiguity, and Language Priors}

Another related line of work studies when VLM outputs are not fully grounded in the current visual evidence. Object hallucination and language-prior benchmarks show that models may produce answers driven by dataset priors or language bias rather than the image~\citep{rohrbach2018object,li2023evaluating,lee2025vlind,vo2025vision}. Broader evaluation suites such as MME~\citep{fu2026mme} also expose perception and cognition failures not captured by standard VQA metrics.

Visual illusion and grounding benchmarks study difficult, ambiguous, or misleading images that can affect model judgments~\citep{guan2024hallusionbench,zhang2023grounding,shahgir2024illusionvqa,zhang2025illusionbench}. Our setting differs in that the target visual state is not shown: the model must infer a rotated outcome from the original image. Related work on visual anagrams, ambigram generation, and multi-view optical illusions shows that visual meaning can change with viewing orientation~\citep{shirakawa2023ambigram,geng2024visual}. These works primarily focus on generating orientation-dependent images, whereas our goal is to evaluate whether VLMs can infer an unobserved rotated outcome from the original image.
\section{Methodology}
\label{sec:methodology}

\subsection{Task Definition}
\label{sec:task_definition}

We study \textit{Rotated-Outcome Prediction}: whether VLMs can infer what would be seen or read after a 180$^\circ$ in-plane rotation without directly observing the rotated target. Given an original image $I$, we define $R_{180}(I)$ as the image obtained by rotating $I$ by 180$^\circ$ within the image plane. The model is shown only $I$ and is asked to answer what would be seen or read in the rotated view $R_{180}(I)$.

This task differs from standard rotation robustness evaluation, where the rotated image is directly provided to the model. In our setting, the model must infer the answer associated with an unobserved rotated visual state. This allows us to separate direct recognition of a rotated input from prediction of the rotated outcome based on the original image.

We focus on 180$^\circ$ rotation because it often yields stable, verifiable semantic changes in visual text and orientation-sensitive images, allowing us to compare endpoint recognition with unseen outcome prediction in a controlled setting.

\subsection{\textsc{RotOutBench} Construction}
\label{sec:benchmark_construction}

We construct \textsc{RotOutBench}, a paired diagnostic benchmark with two subsets: Visual-Rot and TextImage-Rot. Each base example is a rotation pair consisting of an original view and its canonical 180$^\circ$ rotated view. This paired design supports three conditions on the same underlying example: direct reading of the original view, direct reading of the rotated view, and prediction of the rotated target from the original view.

Visual-Rot is built from publicly available web images whose interpretation may change after a 180$^\circ$ rotation. Candidate images include inversion-dependent illustrations, ambigram-like graphics, directional icons, and dual-interpretation images. We manually filter candidates to remove low-quality, overly ambiguous, visually unsupported, or answer-leaking cases. Each retained example is reviewed by two annotators to ensure that the original view, rotated view, question, reference answer, and acceptable keywords are mutually consistent. Disagreements or uncertain cases are resolved by discussion, and examples without a stable rotated interpretation are removed. This process yields 68 paired examples, or 136 image views. We also construct 68 four-way rotation matching questions from these paired examples.

Because Visual-Rot uses open-ended questions, we use manually verified acceptable keywords to reduce evaluation ambiguity. These keywords cover reasonable synonyms, visually equivalent names, and specific category-level descriptions, while excluding overly broad or visually unsupported answers. Thus, Visual-Rot is intended as an open visual diagnostic rather than a fully automatic large-scale leaderboard.

TextImage-Rot provides a more controlled counterpart. It contains 342 paired string examples, yielding 684 image views. Each example has an original rendered string image and a canonical rotated target image. The reference answer is produced by a deterministic 180$^\circ$ rotation rule with character mapping and sequence reversal. Characters without well-defined rotated readings are excluded. The current split balances strings of length 1--5: all single-character and two-character combinations are included, while longer strings are balanced-sampled to avoid the evaluation being dominated by combinatorially many long strings. This keeps the split compact, controlled, and extensible to more characters, fonts, styles, and string lengths.

Together, the two subsets trade off openness and control. Visual-Rot tests whether the phenomenon appears in natural and icon-like images, while TextImage-Rot isolates the recognition-versus-prediction distinction under programmatically generated references. Figure~\ref{fig:benchmark_answer_examples} illustrates the evaluation format; additional construction and evaluation details are provided in Appendix~\ref{app:evaluation_details}.

\subsection{Metrics and Evaluation Protocol}
\label{sec:metrics}

We compute reading and prediction scores as empirical accuracies over paired examples. For model outputs $\hat{Y}$ and reference answers $Y$, we define:
\begin{equation}
\mathrm{Acc}(\hat{Y}, Y) =
N^{-1}\sum_{i=1}^{N}\mathbb{I}[\hat{y}^{(i)} = y^{(i)}],
\label{eq:acc}
\end{equation}
where $N$ is the number of paired examples in the corresponding subset.

We denote model outputs as $\hat{Y}^{Q}_{X}$, where $X$ specifies the input shown to the model and $Q$ specifies the question type. We use $I_{\mathrm{orig}}$ for the original image, $I_{\mathrm{rot}}$ for the rotated target image, $Q_d$ for direct-reading questions, and $Q_p$ for predicted-rotation questions. Exact prompt templates and the image-specific Visual-Rot question variants are described in Appendix~\ref{app:evaluation_details}. We report three main scores:
\begin{align}
R_{\text{orig}} &= \mathrm{Acc}(\hat{Y}^{Q_d}_{I_{\mathrm{orig}}}, Y_{\text{orig}}), \\
R_{\text{rot}}  &= \mathrm{Acc}(\hat{Y}^{Q_d}_{I_{\mathrm{rot}}}, Y_{\text{rot}}), \\
R_{\text{pred}} &= \mathrm{Acc}(\hat{Y}^{Q_p}_{I_{\mathrm{orig}}}, Y_{\text{rot}}).
\end{align}

We also report $\mathrm{Gap}=R_{\text{rot}}-R_{\text{pred}}$, which measures how much easier it is to recognize the rotated target when shown than to infer it from the original view.

For Visual-Rot and TextImage-Rot, $R_{\text{orig}}$, $R_{\text{rot}}$, and $R_{\text{pred}}$ are computed over paired examples. Thus, the main reported scores are computed over 68 Visual-Rot pairs and 342 TextImage-Rot pairs. The auxiliary rotation matching task is evaluated over 68 four-way matching questions constructed from the paired Visual-Rot image views.

For Visual-Rot, open-ended answers are evaluated by normalized keyword matching against manually verified references and acceptable keywords. This evaluation is designed to allow natural answer variation while rejecting overly broad or visually unsupported responses. For TextImage-Rot, outputs are evaluated by normalized string matching against programmatically generated references. Additional details on dataset size, prompt templates, answer parsing, keyword matching, TextImage-Rot generation, and rotation matching are provided in Appendix~\ref{app:evaluation_details}.

\section{Behavioral Results}
\label{sec:behavior}

\begin{figure}[t]
\centering
\includegraphics[width=\columnwidth]{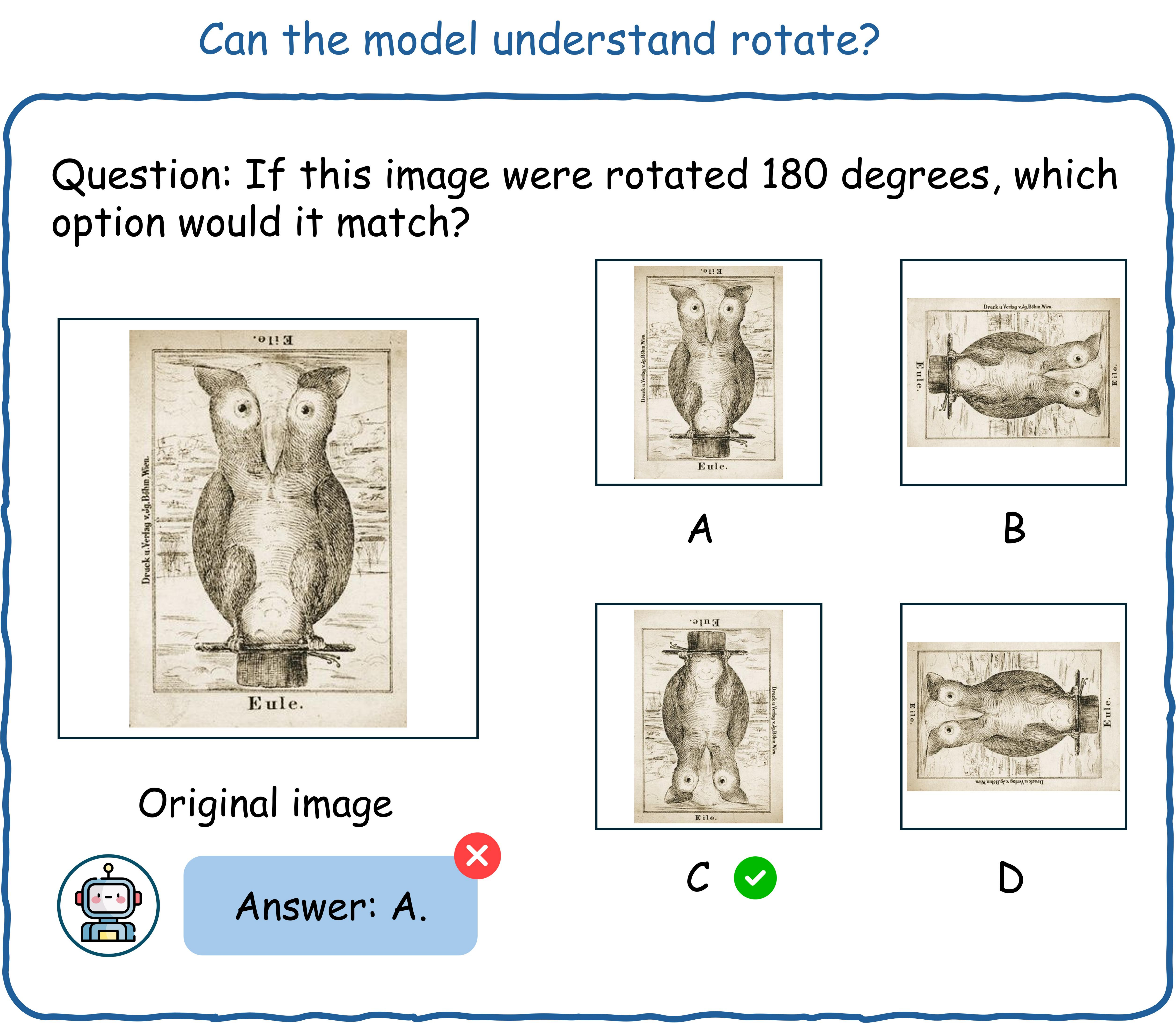}
\caption{Example of the four-way rotation matching task in Visual-Rot. The model is given a query image and four candidates, and must select the candidate corresponding to its 180$^\circ$ rotated view. We use this task as an auxiliary diagnostic.}
\label{fig:rotation_matching_example}
\end{figure}

\subsection{Experimental Setup}
\label{sec:exp_setup}

We evaluate nine representative VLMs on \textsc{RotOutBench}: Qwen2.5-VL-3B~\citep{bai2025qwen25vltechnicalreport}, Qwen3-VL-8B~\citep{bai2025qwen3}, Qwen3.5-VL-9B~\citep{qwen3.5}, Gemma3-4B~\citep{gemmateam2025gemma3technicalreport}, Kimi-VL-A3B~\citep{team2025kimi}, MiniCPM-V-2.6~\citep{yao2024minicpm}, MiniCPM-V-4.5~\citep{yu2025minicpmv45cookingefficient}, InternVL3.5-8B~\citep{wang2025internvl3}, and LLaVA-1.5-7B~\citep{liu2024improved}. All models are evaluated with the protocol defined in Section~\ref{sec:methodology}. Table~\ref{tab:behavior_results} reports direct reading, predicted rotation, the resulting gap, and the auxiliary four-way rotation matching score.

\begin{table*}[t]
\centering
\scriptsize
\resizebox{\textwidth}{!}{%
\begin{tabular}{@{}lrrrr rr rrrr@{}}
\toprule
\multirow{2}{*}{Model} &
\multicolumn{4}{c}{Visual-Rot} &
\multicolumn{2}{c}{Rotation Matching} &
\multicolumn{4}{c}{TextImage-Rot} \\
\cmidrule(lr){2-5}\cmidrule(lr){6-7}\cmidrule(l){8-11}
& $R_{\text{orig}}$ & $R_{\text{rot}}$ & $R_{\text{pred}}$ & Gap
& Acc. & $\Delta_{\mathrm{rand}}$
& $R_{\text{orig}}$ & $R_{\text{rot}}$ & $R_{\text{pred}}$ & Gap \\
\midrule
Qwen2.5-VL-3B & 55.88 & 41.18 & \zerocell{8.82} & \gapcell{32.36} & 23.53 & -1.47 & \textbf{99.42} & \textbf{99.42} & \zerocell{0.29} & \gapcell{99.12} \\
Qwen3-VL-8B & 72.06 & \textbf{58.82} & \zerocell{17.65} & \gapcell{41.17} & 25.00 & 0.00 & 93.27 & 94.74 & \zerocell{2.05} & \gapcell{92.69} \\
Qwen3.5-VL-9B & 73.53 & 57.35 & \zerocell{25.00} & \gapcell{32.35} & 25.00 & 0.00 & 96.20 & 96.20 & \zerocell{2.34} & \gapcell{93.86} \\
Gemma3-4B & 55.88 & 42.65 & \zerocell{14.71} & \gapcell{27.94} & 20.59 & -4.41 & 23.68 & 23.98 & \zerocell{0.29} & \gapcell{23.68} \\
Kimi-VL-A3B & 57.35 & 45.59 & \zerocell{14.71} & \gapcell{30.88} & 25.00 & 0.00 & 98.83 & 99.12 & \zerocell{2.05} & \gapcell{97.08} \\
MiniCPM-V-2.6 & 63.24 & 44.12 & \zerocell{10.29} & \gapcell{33.83} & 23.53 & -1.47 & 97.95 & 97.66 & \zerocell{0.58} & \gapcell{97.08} \\
MiniCPM-V-4.5 & 69.12 & 57.35 & \zerocell{\textbf{26.47}} & \gapcell{30.88} & \textbf{26.47} & 1.47 & 95.32 & 96.49 & \zerocell{\textbf{6.14}} & \gapcell{90.35} \\
InternVL3.5-8B & 66.18 & 47.06 & \zerocell{14.71} & \gapcell{32.35} & 17.65 & -7.35 & 89.18 & 89.77 & \zerocell{2.05} & \gapcell{87.72} \\
LLaVA-1.5-7B & \textbf{75.00} & 57.35 & \zerocell{25.00} & \gapcell{32.35} & 25.00 & 0.00 & 13.16 & 12.57 & \zerocell{1.17} & \gapcell{11.40} \\
\bottomrule
\end{tabular}
}
\caption{Behavioral results on \textsc{RotOutBench}. All values are percentages. $R_{\text{orig}}$ and $R_{\text{rot}}$ measure direct recognition of the original and rotated views, while $R_{\text{pred}}$ measures prediction of the rotated answer from the original view. $\mathrm{Gap}=R_{\text{rot}}-R_{\text{pred}}$ measures the difference between recognizing the rotated target when shown and inferring it from the original view. Main reading and prediction scores are computed over paired examples; Visual-Rot Rotation Matching is evaluated over 68 four-way matching questions constructed from the paired Visual-Rot image views, with $\Delta_{\mathrm{rand}}$ denoting the difference from the 25\% random baseline.}
\label{tab:behavior_results}
\end{table*}

\subsection{Open Visual Rotation}
\label{sec:visual_rot_results}

\paragraph{Open visual cases reveal the prediction challenge.}
Visual-Rot tests rotation-dependent interpretation in open-ended images whose meaning may change after a 180$^\circ$ rotation. As shown in Table~\ref{tab:behavior_results}, $R_{\text{pred}}$ remains low across models, showing that models struggle to infer the rotated outcome from the original image. These results show that rotated-outcome prediction is difficult in open visual cases where rotation changes the apparent object, relation, direction, or semantic interpretation.

\paragraph{Showing candidate images still does not make the correspondence reliable.}
The four-way rotation matching task asks models to select the 180$^\circ$ rotated counterpart from candidate images, as illustrated in Figure~\ref{fig:rotation_matching_example}. Results remain near or below the 25\% random baseline. This provides complementary evidence that the rotation correspondence is difficult even when candidate views are explicitly shown. Since this task evaluates candidate selection rather than open-ended rotated-outcome prediction, we report it as an auxiliary diagnostic alongside the main reading and prediction scores.

\subsection{Controlled Text-Image Rotation}
\label{sec:textimage_results}

\paragraph{TextImage-Rot provides a controlled endpoint comparison.}
TextImage-Rot complements the open visual cases with a controlled setting where each sample has an original string image, a canonical rotated target image, and a programmatically generated rotated answer.

\paragraph{Readable endpoints still lead to near-zero prediction.}
Table~\ref{tab:behavior_results} shows a sharp separation between direct reading and predicted rotation. For models with strong direct-reading accuracy, both $R_{\text{orig}}$ and $R_{\text{rot}}$ are high, confirming that the original string and the rotated target are directly readable. However, $R_{\text{pred}}$ remains near zero, with the best model reaching only 6.14\%.

\paragraph{Simple prompt variants do not remove the failure.}
We rerun the TextImage-Rot predicted-rotation condition with prompt variants that clarify the input view, discourage copying, make the transformation stepwise, or provide a high-level rotation rule. Predicted-rotation accuracy remains near zero across models, and alternative prompts do not produce a consistent improvement. These results suggest that the failure is not merely an artifact of a single prompt wording. Full prompts and results are provided in Appendix~\ref{app:prompt_robustness}.

Figure~\ref{fig:textimage_rotation_example} shows the same pattern qualitatively: the model reads visible strings correctly, but fails to predict what would be read after a 180$^\circ$ rotation.

\begin{figure}[t]
\centering
\includegraphics[width=\columnwidth]{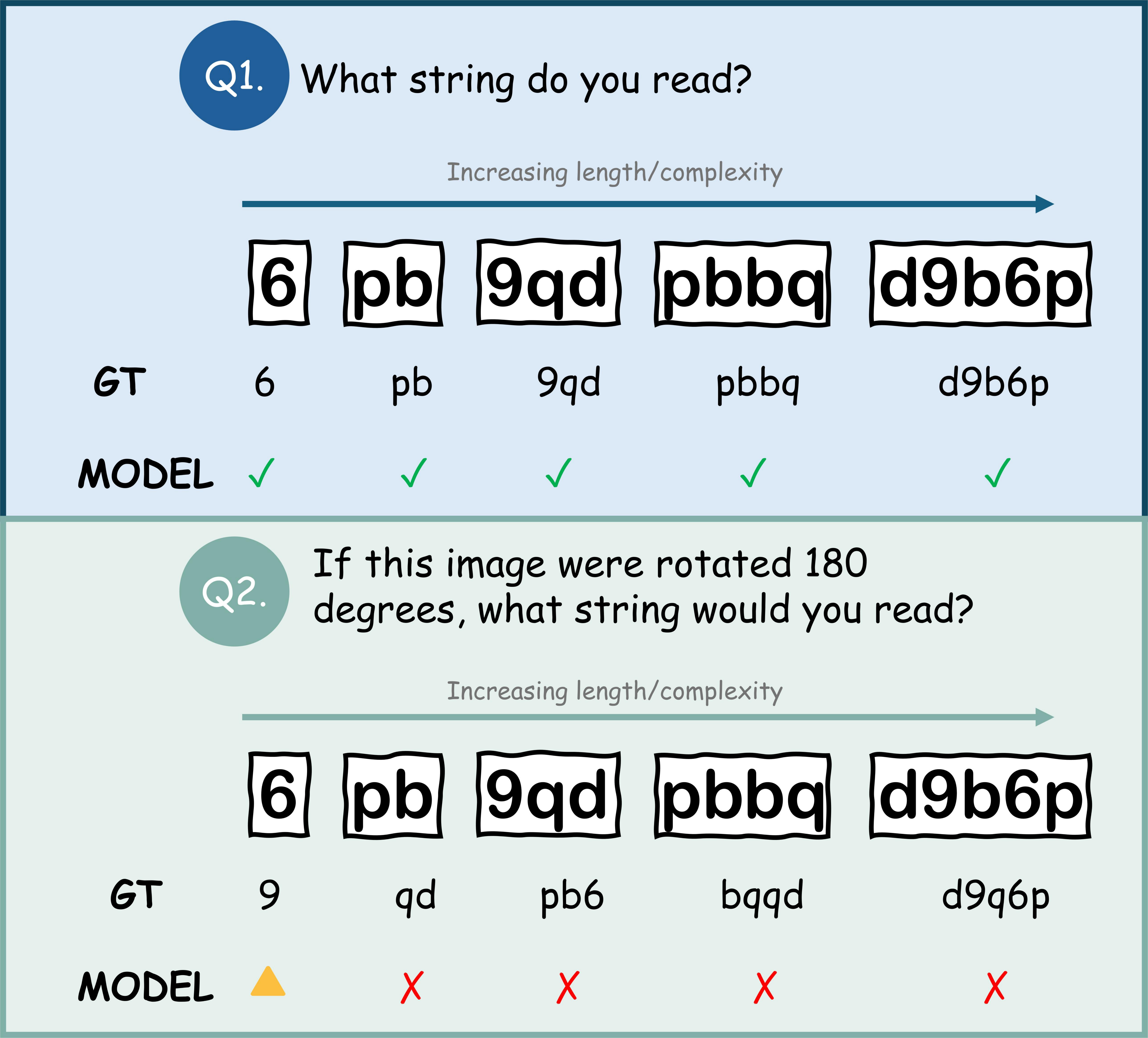}
\caption{Qualitative example from TextImage-Rot. The model reads the relevant string when directly given either the original or rotated image, but fails to infer the rotated reading from the original image alone.}
\label{fig:textimage_rotation_example}
\end{figure}

\begin{figure*}[t]
\centering
\includegraphics[width=0.82\textwidth]{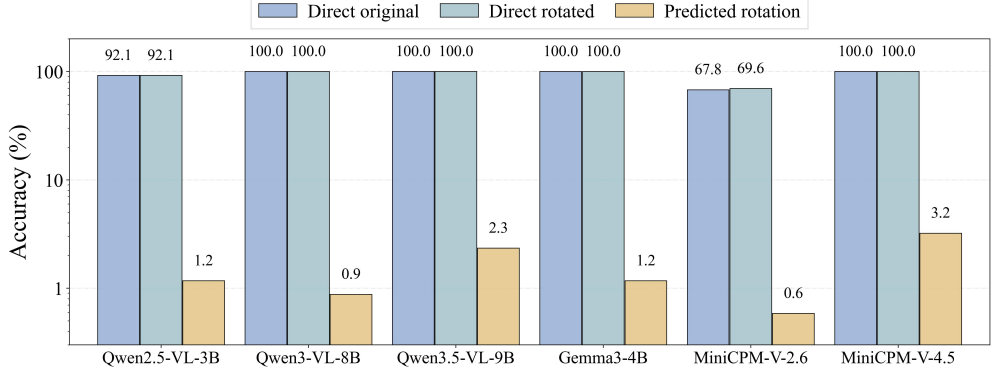}
\caption{Text-only control results across models that support text-only prompts.
Direct reading of the original or rotated string remains easy, while predicting the rotated form from the original string remains difficult. This shows that the failure is not only caused by image rendering or visual-token processing.}
\label{fig:textonly_direct_imagined}
\end{figure*}

\paragraph{High direct reading does not imply prediction.}
The key pattern is clear among models that can directly read both endpoints. On TextImage-Rot, models including Qwen2.5-VL-3B, Qwen3-VL-8B, Qwen3.5-VL-9B, Kimi-VL-A3B, MiniCPM-V-2.6, and MiniCPM-V-4.5 achieve high direct-reading accuracy on both the original and rotated images, yet their predicted-rotation accuracy remains near zero. This rules out a simple endpoint-recognition explanation. Together, the results show that Rotated-Outcome Prediction is not reducible to recognizing a rotated input: models can recognize both endpoints when shown, yet fail to predict the rotated endpoint from the starting image.

\section{Internal Diagnostics}
\label{sec:mechanism}

The behavioral results show a sharp disconnect: models can often read visible rotated targets, but fail to infer the same targets from the original image. We next ask whether this failure is only a surface-level issue in reading rendered text, or whether it also appears during answer formation.

This section provides two levels of evidence. First, a text-only control tests whether the same direct-reading versus prediction separation remains after removing image rendering and visual-token processing. Second, we use Qwen3.5-VL-9B as a case study and analyze hidden states, layer-wise readout, and attention for a recognition-capable model that still fails to predict the rotated outcome.

\subsection{Text-Only Control}
\label{sec:textonly_control}

\paragraph{The failure persists even when images are removed.}
We first test whether the same pattern appears without image input. For models that support text-only prompts, we use the same underlying strings as TextImage-Rot and ask the model either to directly read the given string or to predict what the string would become after a 180$^\circ$ rotation. This mirrors the visual setting, but replaces image inputs $I_{\mathrm{orig}}$ and $I_{\mathrm{rot}}$ with text inputs $T_{\mathrm{orig}}$ and $T_{\mathrm{rot}}$.

As shown in Figure~\ref{fig:textonly_direct_imagined}, direct reading remains much easier than predicted rotation. This indicates that the failure is not only caused by image rendering or visual-token processing. Even when the input is given as text, the transformation itself remains unstable, especially for multi-character strings. Detailed numerical results are provided in Appendix~\ref{app:additional_behavior}.

\subsection{Hidden-State and Readout Analysis}
\label{sec:hidden_readout}

\begin{figure*}[t]
\centering
\includegraphics[width=\textwidth]{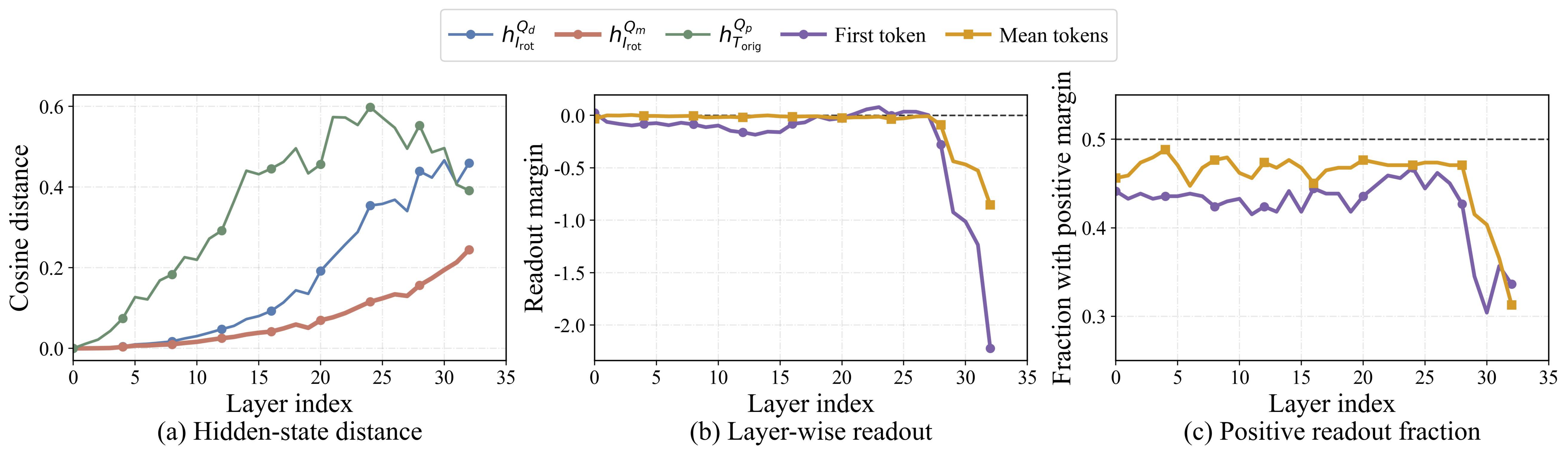}
\caption{Hidden-state and readout analysis on Qwen3.5-VL-9B.
(a) We compare the image-based prediction state with three diagnostic controls: direct rotated-image reading ($h^{Q_d}_{I_{\mathrm{rot}}}$), prompt-matched rotated-image reading ($h^{Q_m}_{I_{\mathrm{rot}}}$), and text-only predicted rotation ($h^{Q_p}_{T_{\mathrm{orig}}}$). Lower distance means greater hidden-state similarity.
(b) Layer-wise readout margin between the rotated target and the original string. Positive values favor the rotated target; negative values favor the original string. We report both first-token and mean-token margins.
(c) Fraction of samples with positive readout margin; the dashed line marks 0.5. The rotated target is not favored for a stable majority of samples. Detailed notation and computation are provided in Appendix~\ref{app:internal_diagnostics}.}
\label{fig:representation_readout_mismatch}
\end{figure*}

\begin{figure*}[t]
\centering
\includegraphics[width=\textwidth]{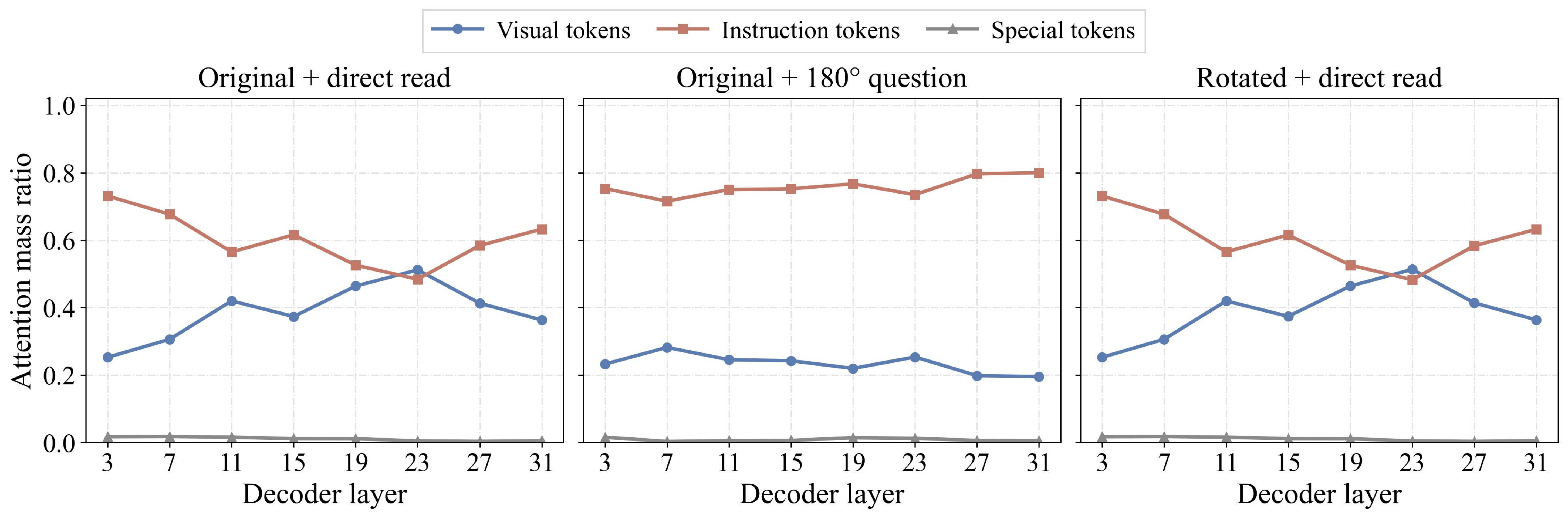}
\caption{Attention allocation across token groups in Qwen3.5-VL-9B. Compared with direct reading, predicted rotation allocates less attention to visual tokens and more attention to instruction tokens, suggesting that the answer state becomes less visually anchored when the rotated outcome must be inferred rather than read directly.}
\label{fig:attention_allocation}
\end{figure*}

\paragraph{The model approaches a rotated-view state, but does not read it out.}
We then examine whether the image-based prediction condition contains rotation-related internal information. The key question is not only whether the model forms a state similar to seeing the rotated target, but whether that state is selected when the final answer is produced.

We compare the image-based prediction state with three diagnostic controls. 
The first is direct reading of the rotated image, denoted as $h^{Q_d}_{I_{\mathrm{rot}}}$, where the rotated target is visible and the model simply reads it. 
The second is prompt-matched reading of the rotated image, denoted as $h^{Q_m}_{I_{\mathrm{rot}}}$, where the rotated target is visible but the prompt still contains rotation-related wording. 
The third is text-only predicted rotation, denoted as $h^{Q_p}_{T_{\mathrm{orig}}}$, where the model receives the original string as text and predicts its rotated form. 
Together, these controls separate different explanations: the image controls test whether the prediction state approaches a state associated with a visible rotated target, while the text-only control tests whether it resembles a text-only transformation path. 
Full notation is provided in Appendix~\ref{app:internal_diagnostics}.

Figure~\ref{fig:representation_readout_mismatch}(a) shows the cosine distance from the image-based prediction state to each control across layers. Lower distance means greater hidden-state similarity. The prediction state is closest to the prompt-matched rotated-image control, rather than to the text-only prediction control. This suggests that the model does not simply behave like a text-only transformation path; its internal state partially approaches the state associated with directly reading the rotated target.

However, this internal similarity does not translate into the final answer. We project the layer-wise hidden state to the vocabulary space and compare the readout score of the rotated target with that of the original string. In Figure~\ref{fig:representation_readout_mismatch}(b), the purple curve uses the first answer token and the yellow curve averages over answer tokens; positive margin favors the rotated target, while negative margin favors the original string. Both readout views shift toward the original string in late layers. Figure~\ref{fig:representation_readout_mismatch}(c) further reports the fraction of samples whose readout margin is positive, with the dashed line marking a majority threshold. The rotated target is not favored for a stable majority of samples. These results suggest that rotation-related information may appear in the prediction state, but the final answer still tends to fall back to the original string.

\subsection{Attention Allocation}
\label{sec:attention_allocation}

\paragraph{Predicted rotation becomes less visually anchored.}
Finally, we compare attention allocation across original-image direct reading, original-image predicted rotation, and rotated-image direct reading. As shown in Figure~\ref{fig:attention_allocation}, predicted rotation allocates less attention mass to visual tokens and more to instruction tokens than the two direct-reading settings. This does not by itself prove that the model ignores the image, but it is consistent with the readout analysis: when the model is asked to predict an unobserved rotated outcome, the answer state becomes less visually anchored than in direct reading.

\paragraph{Summary of the diagnostic evidence.}
Overall, the diagnostics suggest that predicted-rotation failure is not simply caused by the absence of rotation-related information. In Qwen3.5-VL-9B, the hidden state can partially approach a rotated-image reading state, while the final readout still shifts toward the original string. This suggests a late-stage mismatch between transformation-related representation and answer formation.
\section{Conclusion}
\label{sec:conclusion}

We studied \textit{Rotated-Outcome Prediction}, which asks whether VLMs can infer the visual outcome of a 180$^\circ$ rotation without seeing the rotated target. With \textsc{RotOutBench}, we separate reading the original view, reading the rotated endpoint, and predicting that endpoint from the original view. Our results reveal a clear disconnect: many models read both visible endpoints, but not the unseen rotated one. Internal diagnostics suggest that rotation-related information may appear in the prediction state, but is not reliably selected during final answer formation. Overall, evaluating VLMs only on visible transformed inputs can overestimate their transformation reasoning ability. Future evaluations should distinguish visible recognition from inferred visual outcomes.
\section{Limitations}
\label{sec:limitations}

This work focuses on 180$^\circ$ in-plane rotation. We choose this setting because it provides stable and verifiable transformed endpoints for visual text and orientation-sensitive images, allowing us to cleanly separate endpoint recognition from unseen outcome prediction. The current benchmark is therefore not intended to cover all spatial transformations. Future work can extend the same formulation to other rotations, viewpoint changes, deformation, and 3D transformations.

\textsc{RotOutBench} is designed as a focused diagnostic benchmark rather than a large-scale leaderboard. Visual-Rot is intentionally selective because stable 180$^\circ$ semantic reinterpretations are rare and require manual validation, while TextImage-Rot provides a controlled counterpart with deterministic references. The benchmark can be expanded with more fonts, rendering styles, character sets, languages, and transformation families.

Our internal analysis is a model-level case study on Qwen3.5-VL-9B. It illustrates how the failure can appear in hidden states, layer-wise readout, and attention, but does not claim a universal mechanism shared by all VLMs. Extending this analysis to more model families would help determine whether similar internal patterns recur across architectures.

\paragraph{Societal impact.}
This work is a diagnostic study and does not introduce a deployed system. Its potential positive impact is to identify a failure mode in applications involving rotated or non-canonical visual inputs, such as document processing, assistive technologies, and embodied systems. The main risk is over-reliance on VLM inference for orientation-sensitive content without explicit verification.


\bibliography{custom}

\appendix

\section{\textsc{RotOutBench} Evaluation and Dataset Details}
\label{app:evaluation_details}

\subsection{Dataset Size}
\label{app:dataset_size}

Table~\ref{tab:dataset_size} summarizes the scale of \textsc{RotOutBench}. A paired example contains an original view and its canonical 180$^\circ$ rotated view. Image views count both original and rotated views. Rotation matching is reported separately because it evaluates candidate selection rather than open-ended answering.

\begin{table*}[t]
\centering
\small
\begin{tabular}{lrrr}
\toprule
Dataset & Paired examples & Image views & Matching questions \\
\midrule
Visual-Rot & 68 & 136 & 68 \\
TextImage-Rot & 342 & 684 & -- \\
\bottomrule
\end{tabular}
\caption{\textsc{RotOutBench} dataset size. A paired example contains an original view and its canonical 180$^\circ$ rotated view. Image views count both original and rotated views. Rotation matching is an auxiliary task with 68 four-way questions constructed from the paired Visual-Rot image views.}
\label{tab:dataset_size}
\end{table*}

\subsection{Input Conditions and Prompt Templates}
\label{app:prompt_templates}

We use $X$ to denote the input shown to the model and $Q$ to denote the question type. The question label alone does not determine the input. For example, $Q_d$ only denotes a direct-reading question, and the input can be either the original image $I_{\mathrm{orig}}$, the rotated target image $I_{\mathrm{rot}}$, the original text string $T_{\mathrm{orig}}$, or the rotated text string $T_{\mathrm{rot}}$. A complete evaluation condition is therefore specified by the pair $(X,Q)$, or equivalently by outputs such as $\hat{Y}^{Q_d}_{I_{\mathrm{orig}}}$ and $\hat{Y}^{Q_p}_{I_{\mathrm{orig}}}$.

For TextImage-Rot, we use fixed prompt templates. Direct reading uses $Q_d$:
\begin{quote}
What string do you read? Reply with exactly the string.
\end{quote}
This prompt is used with either the original text image $I_{\mathrm{orig}}$ or the rotated target image $I_{\mathrm{rot}}$.

Predicted rotation uses $Q_p$:
\begin{quote}
If this image were rotated 180 degrees, what string would you read? Reply with exactly the string.
\end{quote}
This prompt is used with the original text image $I_{\mathrm{orig}}$, and the reference answer is the rotated target string.

For the text-only control, we use the same question types but replace image inputs with text inputs. Direct reading uses:
\begin{quote}
What string do you read? Reply with exactly the string.
\end{quote}
Predicted rotation uses:
\begin{quote}
If this string were rotated 180 degrees, what string would you read? Reply with exactly the string.
\end{quote}

For Visual-Rot, the exact wording is image-specific because different images require different semantic targets. We therefore use $Q_d$ and $Q_p$ only as condition labels. A Visual-Rot direct-reading question asks about the target visible in the shown image, while a Visual-Rot predicted-rotation question asks about the corresponding target after a 180$^\circ$ rotation without revealing the rotated answer.

For the four-way rotation matching task, we use:
\begin{quote}
If this image were rotated 180 degrees, which option would match? Answer with A, B, C, or D.
\end{quote}

For the prompt-matched rotated-image control used in the internal analysis, we use $Q_m$:
\begin{quote}
This image has already been rotated 180 degrees. What string do you read now? Reply with exactly the string.
\end{quote}
This condition is paired with the rotated target image $I_{\mathrm{rot}}$. It keeps rotation-related wording while making the rotated target directly visible.

For models that produce explanatory text despite the instruction, we parse the final answer using the normalization rules described below.

\subsection{Answer Parsing and Normalization}
\label{app:parsing_normalization}

For Visual-Rot, model outputs are open-ended. We lowercase the output, remove punctuation, and normalize minor formatting differences. A response is counted as correct if it matches the reference answer or one of the manually verified acceptable keywords.

The acceptable keywords are used to handle natural variation in free-form answers. They may include synonyms, visually equivalent names, common alternative names, or reasonable category-level descriptions. However, we exclude overly broad terms, visually unsupported answers, and semantically different interpretations. This prevents the evaluation from becoming overly permissive while still allowing natural variation in open-ended responses.

For TextImage-Rot, the expected answer is a specific character or string. We therefore use normalized string matching against the programmatically generated target answer. We remove surrounding whitespace and punctuation introduced by the model, but do not use semantic keyword matching for this subset.

For rotation matching, we parse the model output as an option letter. If the response contains multiple option letters, we use the first valid option letter unless the answer explicitly states a different final choice. Responses without a valid option letter are counted as incorrect.

\subsection{Visual-Rot Review and Keyword Annotation}
\label{app:visual_rot_review}

Each Visual-Rot example is manually reviewed to ensure that the rotated interpretation is visually grounded, that the question can be answered from the image, and that the reference answer captures the intended rotated-view interpretation. Samples are removed if the rotated view remains too ambiguous, if multiple incompatible interpretations are equally plausible, or if the answer depends on information not visible in the image.

For each retained example, we check the consistency among the original image, rotated image, question, reference answer, and acceptable keywords. The acceptable keywords are manually reviewed to include reasonable variations while excluding overly generic or visually unsupported descriptions. For example, a rotated target that can reasonably be described with a specific object name may also allow a common synonym or a visually justified category-level description, but not an unrelated object name or a generic term that does not preserve the intended visual meaning.

\subsection{TextImage-Rot Generation}
\label{app:textimage_generation}

TextImage-Rot is generated from predefined character sets, string templates, and font styles. Each source string has a deterministic 180$^\circ$ rotated target. The target is produced by applying a character-level rotation mapping and reversing the character order.

For a source string
\[
s = c_1 c_2 \cdots c_n,
\]
its rotated target is defined as
\[
\mathrm{rot180}(s)
=
\rho(c_n)\rho(c_{n-1})\cdots\rho(c_1),
\]
where $\rho(\cdot)$ maps each character to its 180$^\circ$ rotated counterpart. Characters without a well-defined rotated counterpart are excluded.

We verify the consistency among the source string, rendered original image, rendered rotated image, and reference answer using scripts, followed by manual inspection. This ensures that the original image, rotated target image, and programmatic answer correspond to the same transformation rule.

\subsection{TextImage-Rot String Distribution}
\label{app:textimage_distribution}

TextImage-Rot uses strings of length 1 to 5. We include all available single-character and two-character combinations, and balance-sample longer strings to keep the dataset compact while avoiding dominance by any single length.

\begin{table}[t]
\centering
\small
\begin{tabular}{rrl}
\toprule
Length & Count & Generation strategy \\
\midrule
1 & 6   & all single characters \\
2 & 36  & all $6^2$ combinations \\
3 & 100 & balanced sampling \\
4 & 100 & balanced sampling \\
5 & 100 & balanced sampling \\
\midrule
Total & 342 & $6+36+100+100+100$ \\
\bottomrule
\end{tabular}
\caption{String-length distribution in TextImage-Rot. Longer strings are balance-sampled to avoid combinatorial dominance while keeping the split compact and extensible.}
\label{tab:textimage_length_distribution}
\end{table}

\subsection{Rotation Matching Task}
\label{app:rotation_matching_task}

The four-way rotation matching task is used as an auxiliary diagnostic. It is evaluated over 68 four-way questions constructed from the paired Visual-Rot image views. Each question contains one query image and four candidate images. One candidate is the corresponding 180$^\circ$ rotated counterpart, while the other candidates are distractors. Candidate order is randomized.

We report the matching accuracy and its difference from the random baseline:
\begin{equation}
\Delta_{\mathrm{rand}}
=
\mathrm{Acc}_{\mathrm{match}} - 25.
\end{equation}

Since the task has four options, the random baseline is 25\%. We report this task as an auxiliary diagnostic because it evaluates candidate selection in addition to rotated-outcome prediction. It therefore complements the open-ended prediction scores rather than serving as the main evidence.

\section{Additional Behavioral Results}
\label{app:additional_behavior}

\subsection{Text-Only Control Results}
\label{app:textonly_results}

The text-only control mirrors the TextImage-Rot protocol, but replaces image inputs with plain-text strings. This control asks whether the direct-reading versus prediction separation remains after removing image rendering and visual-token processing.

\begin{table}[t]
\centering
\small
\begin{tabular}{lrrrr}
\toprule
Model
& $R^{\mathrm{text}}_{\mathrm{orig}}$
& $R^{\mathrm{text}}_{\mathrm{rot}}$
& $R^{\mathrm{text}}_{\mathrm{pred}}$
& $G^{\mathrm{text}}$ \\
\midrule
Qwen2.5-VL-3B   & 92.11  & 92.11  & \zerocell{1.17} & \gapcell{90.94} \\
Qwen3-VL-8B     & \textbf{100.00} & \textbf{100.00} & \zerocell{0.88} & \gapcell{99.12} \\
Qwen3.5-VL-9B   & \textbf{100.00} & \textbf{100.00} & \zerocell{2.34} & \gapcell{97.66} \\
Gemma3-4B       & \textbf{100.00} & \textbf{100.00} & \zerocell{1.17} & \gapcell{98.83} \\
MiniCPM-V-2.6   & 67.84  & 69.59  & \zerocell{0.58} & \gapcell{69.01} \\
MiniCPM-V-4.5   & \textbf{100.00} & \textbf{100.00} & \zerocell{\textbf{3.22}} & \gapcell{96.78} \\
\bottomrule
\end{tabular}
\caption{Text-only control results. All values are percentages. $R^{\mathrm{text}}_{\mathrm{pred}}$ is highlighted because predicted rotation remains close to zero even when the input is provided as plain text. The control is evaluated on models that support text-only prompts in our setup.}
\label{tab:textonly_results}
\end{table}

Let $T_{\mathrm{orig}}$ denote the original source string, and let $T_{\mathrm{rot}}$ denote the corresponding 180$^\circ$ rotated target string. This is the text analogue of the visual setting, where $I_{\mathrm{orig}}$ and $I_{\mathrm{rot}}$ denote the original image and the canonical rotated target image. We use $Q_d$ for direct-reading questions and $Q_p$ for predicted-rotation questions:
\begin{align}
R^{\mathrm{text}}_{\mathrm{orig}}
&=
\mathrm{Acc}(\hat{Y}^{Q_d}_{T_{\mathrm{orig}}}, Y_{\mathrm{orig}}), \\
R^{\mathrm{text}}_{\mathrm{rot}}
&=
\mathrm{Acc}(\hat{Y}^{Q_d}_{T_{\mathrm{rot}}}, Y_{\mathrm{rot}}), \\
R^{\mathrm{text}}_{\mathrm{pred}}
&=
\mathrm{Acc}(\hat{Y}^{Q_p}_{T_{\mathrm{orig}}}, Y_{\mathrm{rot}}).
\end{align}

We also report the text-only gap:
\begin{equation}
G^{\mathrm{text}}
=
R^{\mathrm{text}}_{\mathrm{rot}}
-
R^{\mathrm{text}}_{\mathrm{pred}}.
\end{equation}

The results show that direct reading remains high for most models, while predicted rotation remains close to zero. This rules out a purely image-level explanation: the failure is not only caused by image rendering or visual-token processing. Instead, even when the original string is given explicitly as text, models still struggle to apply the 180$^\circ$ transformation and produce the rotated target. This supports the main behavioral result that the difficulty lies in deriving an unseen rotated outcome, not merely in recognizing a visible endpoint.

\subsection{Prompt Robustness}
\label{app:prompt_robustness}

We test whether the low predicted-rotation accuracy on TextImage-Rot is caused by a single underspecified prompt. We evaluate five prompts for the predicted-rotation condition.

\textbf{P0 (Base):} ``If this image were rotated 180 degrees, what string would you read? Reply with exactly the string.'' This is the base prompt used in the main experiments.

\textbf{P1 (Explicit):} ``The image shown is the original image, not the rotated one. If it were rotated 180 degrees, what string would be visible? Reply with only the rotated string.''

\textbf{P2 (Anti-copy):} ``Do not copy the string currently visible. Predict the string after the whole image is rotated 180 degrees. Reply with only the rotated string.''

\textbf{P3 (Stepwise):} ``First read the current string internally, then apply a 180-degree image rotation, and output only the final rotated string.''

\textbf{P4 (Rule-aware):} ``For a 180-degree image rotation, the character order is reversed and each character appears as its rotated counterpart. Apply this rule to the string in the image. Reply with only the rotated string.'' P4 additionally provides a high-level rotation rule and is reported as a rule-aware variant.

P1--P3 test instruction-level wording changes, while P4 gives an additional high-level rotation rule.

\begin{table}[t]
\centering
\footnotesize
\resizebox{\columnwidth}{!}{%
\setlength{\tabcolsep}{4pt}
\begin{tabular}{@{}p{2.2cm}rrrrr@{}}
\toprule
Model & P0 & P1 & P2 & P3 & P4 \\
\midrule
Qwen2.5-VL-3B  & 0.29 & 0.00 & 0.00 & 1.75 & 0.00 \\
Qwen3-VL-8B    & 2.05 & 1.75 & 1.46 & 0.29 & 0.00 \\
Qwen3.5-VL-9B  & 2.34 & 2.92 & 2.34 & 2.63 & 2.34 \\
Gemma3-4B      & 0.29 & 0.88 & 0.58 & 0.88 & 0.00 \\
Kimi-VL-A3B    & 2.05 & 0.88 & 1.17 & 0.00 & 0.29 \\
MiniCPM-V-2.6  & 0.58 & 0.88 & 0.88 & 1.46 & 0.88 \\
MiniCPM-V-4.5  & 6.14 & 4.39 & 5.85 & 5.85 & 3.22 \\
InternVL3.5-8B & 2.05 & 2.05 & 1.75 & 1.46 & 1.17 \\
LLaVA-1.5-7B   & 1.17 & 1.17 & 1.17 & 1.17 & 0.58 \\
\bottomrule
\end{tabular}%
}
\caption{Prompt robustness results on TextImage-Rot predicted rotation (\%). P0--P4 denote the prompt variants described above.}
\label{tab:prompt_ablation_results}
\end{table}

Table~\ref{tab:prompt_ablation_results} shows that the failure persists across prompt variants. To further illustrate the failure patterns, Table~\ref{tab:prompt_ablation_failures} shows one representative failure example per string length. For each example, we report the model's raw output under each prompt variant, truncated to the first 20 characters. The source string and target are shown in lowercase, matching the actual image content. ``--'' indicates an empty output.

\begin{table*}[t]
\centering
\scriptsize
\resizebox{\textwidth}{!}{%
\begin{tabular}{@{}lllll@{}}
\toprule
& \multicolumn{4}{c}{Prompt variant} \\
\cmidrule(l){2-5}
Model & P1 & P2 & P3 & P4 \\
\midrule
\multicolumn{5}{@{}l}{\textbf{L=1:} shown ``q'' $\rightarrow$ target ``b''} \\
\midrule
Qwen2.5-VL-3B  & q & q & The letter ``q'' rotat\ldots & q \\
Qwen3-VL-8B    & q & q & q & q \\
Qwen3.5-VL-9B  & q & q & p & q \\
Gemma3-4B      & Q & q & q & q \\
Kimi-VL-A3B    & q & q & q & To solve the problem\ldots \\
MiniCPM-V-2.6  & The rotated string, w\ldots & q & The final rotated strin\ldots & The rotated string, foll\ldots \\
MiniCPM-V-4.5  & dq & q & q & q \\
InternVL3.5-8B & q & q & 1 & 1 \\
LLaVA-1.5-7B   & A & AQ & q & Qq \\
\midrule
\multicolumn{5}{@{}l}{\textbf{L=2:} shown ``6b'' $\rightarrow$ target ``q9''} \\
\midrule
Qwen2.5-VL-3B  & b6 & b6 & The string ``6b'' is a \ldots & 6b \\
Qwen3-VL-8B    & b6 & b6 & 9b6 & b6 \\
Qwen3.5-VL-9B  & b9 & b9 & b9 & b9 \\
Gemma3-4B      & b6 & b6 & b6 & b6 \\
Kimi-VL-A3B    & 6b & 6b & To solve the task, we\ldots & b6 \\
MiniCPM-V-2.6  & b6 & 9b & The final rotated strin\ldots & b6 \\
MiniCPM-V-4.5  & 9q & 9q & 9q & b9 \\
InternVL3.5-8B & 6b & 6b & b6 & 6b \\
LLaVA-1.5-7B   & 6b & 68b & 68b & b68 \\
\midrule
\multicolumn{5}{@{}l}{\textbf{L=3:} shown ``bp6'' $\rightarrow$ target ``9dq''} \\
\midrule
Qwen2.5-VL-3B  & 6p & 6p & The string ``bp6'' rema\ldots & 6pbb \\
Qwen3-VL-8B    & 6pob & 6pob & 6p6 & 6pob \\
Qwen3.5-VL-9B  & 9pqb & 9pqb & 9pqb & 9po \\
Gemma3-4B      & 6p6b & 6p6b & 6p b bp6 & 6p6b \\
Kimi-VL-A3B    & 6pmb & 6pqb & To solve the problem\ldots & 6pb \\
MiniCPM-V-2.6  & ebp6 & 6p9 & The final rotated strin\ldots & ebp6 \\
MiniCPM-V-4.5  & 9d & 9dqb & 9dqb & 6pbb \\
InternVL3.5-8B & 6pb & 6pb & 6pb & 6p6 \\
LLaVA-1.5-7B   & BPP66 & BPP66 & bpp66 & BPP66 \\
\midrule
\multicolumn{5}{@{}l}{\textbf{L=4:} shown ``6dq9'' $\rightarrow$ target ``6bp9''} \\
\midrule
Qwen2.5-VL-3B  & 9q6d & 9q6d6 & The string ``6dq9'' is \ldots & 9q6 \\
Qwen3-VL-8B    & 6qdl9 & 6q9d6 & 9qdl6 & 9qdl6 \\
Qwen3.5-VL-9B  & 6dq9 & 6dq9 & 6dq9 & 6dq9 \\
Gemma3-4B      & 9qgd6 & 9qdp6 & 9dqp6 & 9qdp6 \\
Kimi-VL-A3B    & 6dq9 & 6dq9 & To solve the task, we\ldots & 9q6 \\
MiniCPM-V-2.6  & 9qdd & 9qdd & The final rotated strin\ldots & 9q6d \\
MiniCPM-V-4.5  & 6dq9 & 6dqw9 & 6dq9 & 9qdq6 \\
InternVL3.5-8B & 6dq9 & 6dq9 & 6dq9 & 9qd6 \\
LLaVA-1.5-7B   & DQ96 & DQ96 & Q96DQ96DQ96DQ96DQ96\ldots & DQ96DQ96DQ96DQ96DQ96\ldots \\
\midrule
\multicolumn{5}{@{}l}{\textbf{L=5:} shown ``9dp6b'' $\rightarrow$ target ``q9dp6''} \\
\midrule
Qwen2.5-VL-3B  & b6p9d9 & b6p9d9 & The string ``9dp6b'' is\ldots & 6b9dp \\
Qwen3-VL-8B    & b6p9d & b6p9d9 & b6p9d & b6p9d \\
Qwen3.5-VL-9B  & b9p6d & b9p6d9 & b9p6d9 & b9p6d9 \\
Gemma3-4B      & b6dpd9 & b6pd9g & b6dp9d & b6pd9b \\
Kimi-VL-A3B    & 6bp9d & b6pd9 & To solve the task, we\ldots & b6pd9 \\
MiniCPM-V-2.6  & b6p9 & b9p6d & The string ``9dp6b'' ro\ldots & b6p9 \\
MiniCPM-V-4.5  & b9p6dp & b9p6d6 & b9p6d6 & b9p6dp6 \\
InternVL3.5-8B & b6pd9 & b6pd9 & b6pd9 & b6pd9 \\
LLaVA-1.5-7B   & DP6L & DP66 & DP66L & DP6L \\
\bottomrule
\end{tabular}%
}
\caption{Representative failure examples across prompt variants, organized by string length. Each row shows one model's raw output for the same image under different prompt variants. Raw outputs are truncated to 20 characters. P0 is omitted because it uses the same prompt as the main experiment.}
\label{tab:prompt_ablation_failures}
\end{table*}

\section{Internal Diagnostic Details}
\label{app:internal_diagnostics}

\subsection{Conditions and Notation}
\label{app:conditions_notation}

For the internal analysis, we focus on Qwen3.5-VL-9B and use the TextImage-Rot setting. Let $I_{\mathrm{orig}}$ denote the original text image, $I_{\mathrm{rot}}$ denote the canonical 180$^\circ$ rotated target image, $T_{\mathrm{orig}}$ denote the original source string in text form, and $T_{\mathrm{rot}}$ denote the rotated target string in text form.

As in Appendix~\ref{app:prompt_templates}, a condition is specified by both the input $X$ and the question type $Q$. We use $Q_d$ for direct reading, $Q_p$ for predicted rotation, and $Q_m$ for the prompt-matched rotated-image control. The wording of $Q_p$ is instantiated as ``this image'' for image inputs and ``this string'' for text inputs.

The main image-based prediction condition is:
\[
h^{Q_p}_{I_{\mathrm{orig}}}(l),
\]
where the model receives $I_{\mathrm{orig}}$ and is asked to predict what would be read after a 180$^\circ$ rotation. Here, $h(l)$ denotes the answer-position hidden state at decoder layer $l$.

We compare this condition with three diagnostic controls:
\begin{itemize}
    \item $h^{Q_d}_{I_{\mathrm{rot}}}(l)$: direct reading of the rotated target image.
    \item $h^{Q_m}_{I_{\mathrm{rot}}}(l)$: prompt-matched reading of the rotated target image, where the rotated image is visible but the prompt keeps rotation-related wording.
    \item $h^{Q_p}_{T_{\mathrm{orig}}}(l)$: text-only predicted rotation, where the model receives the original string as text and predicts its rotated form.
\end{itemize}

The prompt-matched control helps distinguish similarity to a visible rotated target from similarity caused only by rotation-related wording.

\subsection{Hidden-State Distance}
\label{app:hidden_distance}

Figure~\ref{fig:representation_readout_mismatch}(a) measures how close the image-based prediction state is to each control. For each decoder layer $l$, we compute cosine distance from $h^{Q_p}_{I_{\mathrm{orig}}}(l)$ to the three controls:
\begin{align}
D_{\mathrm{direct}}(l)
&=
1 -
\cos\left(
h^{Q_p}_{I_{\mathrm{orig}}}(l),
h^{Q_d}_{I_{\mathrm{rot}}}(l)
\right), \\
D_{\mathrm{matched}}(l)
&=
1 -
\cos\left(
h^{Q_p}_{I_{\mathrm{orig}}}(l),
h^{Q_m}_{I_{\mathrm{rot}}}(l)
\right), \\
D_{\mathrm{text}}(l)
&=
1 -
\cos\left(
h^{Q_p}_{I_{\mathrm{orig}}}(l),
h^{Q_p}_{T_{\mathrm{orig}}}(l)
\right).
\end{align}

A smaller distance means that the image-based prediction state is more similar to that control condition. In Figure~\ref{fig:representation_readout_mismatch}(a), the prediction state is closest to the prompt-matched rotated-image control $h^{Q_m}_{I_{\mathrm{rot}}}$. This suggests that the prediction state is closer to a rotated-image reading condition than to a text-only transformation condition.

\subsection{Layer-Wise Readout}
\label{app:layerwise_readout}

Hidden-state similarity alone does not tell us which answer the model is preparing to output. A prediction state may be close to a rotated-image control, but the final answer can still be selected from a different competing representation. We therefore apply the final normalization and language modeling head to the hidden state at each layer, producing a layer-wise readout proxy.

Let $Y_{\mathrm{orig}}$ denote the original string answer and $Y_{\mathrm{rot}}$ denote the rotated target answer. For a candidate answer $Y=(y_1,\ldots,y_K)$, we compute a mean-token readout score:
\begin{equation}
S_{\mathrm{mean}}^{(l)}(Y)
=
\frac{1}{K}
\sum_{k=1}^{K}
z^{(l)}(y_k),
\end{equation}
where $z^{(l)}(y_k)$ is the logit assigned to token $y_k$ by the layer-$l$ readout. We also compute a first-token score:
\begin{equation}
S_{\mathrm{first}}^{(l)}(Y)
=
z^{(l)}(y_1).
\end{equation}

The readout margin compares the rotated target with the original string:
\begin{align}
M_{\mathrm{mean}}^{(l)}
&=
S_{\mathrm{mean}}^{(l)}(Y_{\mathrm{rot}})
-
S_{\mathrm{mean}}^{(l)}(Y_{\mathrm{orig}}), \\
M_{\mathrm{first}}^{(l)}
&=
S_{\mathrm{first}}^{(l)}(Y_{\mathrm{rot}})
-
S_{\mathrm{first}}^{(l)}(Y_{\mathrm{orig}}).
\end{align}

A positive margin means that the layer-wise readout favors the rotated target. A negative margin means that it favors the original string. Figure~\ref{fig:representation_readout_mismatch}(b) shows that the late-layer margin becomes negative, meaning that the readout shifts toward the original string even though the hidden state is closest to the prompt-matched rotated-image control.

\subsection{Positive-Margin Fraction}
\label{app:positive_margin_fraction}

To test whether the rotated target is favored for most samples, we also compute the fraction of samples with a positive readout margin:
\begin{equation}
F^{(l)}
=
\frac{1}{N}
\sum_{i=1}^{N}
\mathbb{I}
\left[
M_i^{(l)} > 0
\right].
\end{equation}

Figure~\ref{fig:representation_readout_mismatch}(c) reports this fraction across layers. A value above $0.5$ means that more than half of the samples favor the rotated target at that layer. The curve does not stay above this threshold, showing that the rotated target is not selected for a stable majority of samples. This sample-level view complements the average margin in Figure~\ref{fig:representation_readout_mismatch}(b): the late-layer shift is not just a mean-effect artifact, but reflects an unstable preference across examples.

\subsection{Attention Grouping}
\label{app:attention_grouping}

For attention allocation, we group input tokens into visual tokens, instruction tokens, and special tokens. At each selected decoder layer, we measure the attention mass from the answer position to each group:
\begin{equation}
A_g^{(l)}
=
\sum_{j \in g}
\alpha^{(l)}_{\mathrm{ans},j},
\end{equation}
where $g$ is a token group and $\alpha^{(l)}_{\mathrm{ans},j}$ is the attention weight from the answer position to token $j$ at layer $l$.

The text-only condition is excluded from this attention comparison because the goal is to compare how image-input conditions distribute attention between visual tokens and instruction tokens. This attention analysis is descriptive. Lower visual-token attention does not by itself prove that the model ignores visual information, but it provides supporting evidence that predicted rotation is less visually anchored than direct reading.

\section{Artifact Use and Implementation}
\label{app:artifact_implementation}

\subsection{Artifact Use and Intended Use}
\label{app:artifact_use}

We use publicly released VLMs for research evaluation and cite their corresponding papers or technical reports. \textsc{RotOutBench} is intended as a diagnostic research benchmark for evaluating rotated-outcome prediction in VLMs. It should not be used as a deployed decision system or as a replacement for human verification in safety-sensitive orientation-dependent applications. For images collected from public sources, benchmark materials should be used and released according to the corresponding access conditions.

\subsection{Implementation Details and Compute Budget}
\label{app:implementation_details}

All experiments are inference-only. We evaluate the released model variants listed in Table~\ref{tab:behavior_results}. For each model, we use its released checkpoint and the corresponding tokenizer and processor from the official or commonly used open-source implementation. For models supported by VLMEvalKit~\citep{duan2024vlmevalkit}, we use it for standardized inference; the remaining models are evaluated with model-specific scripts under the same prompts, decoding settings, and parsing rules. Unless otherwise stated, we use temperature 0 for deterministic short-answer evaluation, with the same prompt templates and parsing rules across models.

The main behavioral evaluation, text-only control, prompt robustness experiments, and internal diagnostic analyses were run on a server with four NVIDIA A100 GPUs. The total compute budget was approximately 20 GPU-hours. The internal diagnostics are conducted on Qwen3.5-VL-9B and require extracting hidden states, layer-wise readout scores, and attention weights under the conditions described in Appendix~\ref{app:internal_diagnostics}.

\end{document}